\documentclass[11pt,a4paper]{article}
\usepackage[hyperref]{emnlp-ijcnlp-2019}
\usepackage{times}
\usepackage{latexsym}

\usepackage{url}
\usepackage{verbatim}

\usepackage{microtype}
\usepackage{xspace}
\usepackage{booktabs}
\usepackage{adjustbox}
\usepackage{multirow}
\usepackage[maxfloats=150]{morefloats}
\usepackage{makecell}
\usepackage{graphicx}
\usepackage{subcaption}
\usepackage[shortlabels]{enumitem}

\usepackage{cleveref}
\crefname{section}{\S}{\S\S}
\Crefname{section}{\S}{\S\S}
\crefname{table}{Table}{}
\crefname{figure}{Figure}{}
\crefname{algorithm}{Algorithm}{}
\crefname{equation}{eq.}{}
\crefname{appendix}{App.}{}
\crefname{ExNo}{Sentence}{}
\crefformat{section}{\S#2#1#3}  % remove space between section symbol and the number

\usepackage{amssymb}

\usepackage{pifont}% http://ctan.org/pkg/pifont
\newcommand{\cmark}{\ding{51}}%
\newcommand{\xmark}{\ding{55}}%

\usepackage{todonotes}
\makeatletter
\newcommand*\iftodonotes{\if@todonotes@disabled\expandafter\@secondoftwo\else\expandafter\@firstoftwo\fi}  % defines \iftodonotes{<true>}{<false>}, thanks to https://tex.stackexchange.com/questions/126559/conditional-based-on-packageoption
\makeatother

\newcommand{\word}[1]{\textit{#1}}
\newcommand{\defn}[1]{\textbf{#1}}
\newcommand{\model}[1]{\textsc{#1}}

\newcommand{\lang}[1]{{#1}}

\newcommand{\lennum}[1]{{\color{gray} #1}}

\newcommand{\B}[1]{\textbf{#1}}
\newcommand{\T}[1]{\textsc{\MakeLowercase{#1}}}
\newcommand\tab[1][1cm]{\hspace*{#1}}

\aclfinalcopy % Uncomment this line for the final submission

\title{Don't Forget the Long Tail! A Comprehensive Analysis of Morphological Generalization in Bilingual Lexicon Induction}

\author{Paula Czarnowska\textsuperscript{1} \tab Sebastian Ruder\textsuperscript{2}\footnotemark $\:$ \tab  Edouard Grave\textsuperscript{3}\\
\textbf{Ryan Cotterell\textsuperscript{1} \tab Ann Copestake\textsuperscript{1}}\\
  \textsuperscript{1}University of Cambridge \tab \textsuperscript{2}National University of Ireland \tab \textsuperscript{3}Facebook AI Research \\
  \texttt{pjc211@cam.ac.uk}  \tab[0.5cm] \texttt{sebastian@ruder.io} \tab[0.5cm] \texttt{egrave@fb.com}\\
  \texttt{rdc42@cam.ac.uk} \tab[0.5cm] \texttt{aac10@cam.ac.uk}
}
\date{}

\begin{document}
\maketitle
\begin{abstract}
  Human translators routinely have to translate rare inflections of words---due to the Zipfian distribution of words in a language.   When translating from Spanish, a good translator would have no problem identifying the proper translation of a statistically rare inflection such as \word{hablar{\'a}mos}. Note the lexeme itself, \word{hablar},
  is relatively common. In this work, we investigate whether state-of-the-art bilingual lexicon inducers are capable of learning this kind of generalization. We introduce 40 \emph{morphologically complete} dictionaries in 10 languages\footnote{The dictionaries are available at \url{https://github.com/pczarnowska/morph_dictionaries}.} and evaluate three of the state-of-the-art models on the task of translation of less frequent morphological forms. We demonstrate that the performance of state-of-the-art models drops considerably when evaluated on infrequent morphological inflections and then show that adding a simple morphological constraint at training time improves the performance, proving that the bilingual lexicon inducers can benefit from better encoding of morphology.\looseness=-1
\end{abstract}

\newenvironment{starfootnotes}
  {\par\edef\savedfootnotenumber{\number\value{footnote}}
  \renewcommand{\thefootnote}{$\star$} 
  \setcounter{footnote}{0}}
  {\par\setcounter{footnote}{\savedfootnotenumber}}
  
\begin{starfootnotes}
\footnotetext{Sebastian is now affiliated with DeepMind.}
\end{starfootnotes}

\section{Introduction}

Human translators exhibit remarkable generalization capabilities and are able to translate even rare inflections they may have never seen before. Indeed, this skill is necessary for translation since language follows a Zipfian distribution \cite{zipf1949human}: a large number of the tokens in a translated text will come from rare types, including rare inflections of common lexemes. For instance, a Spanish translator will most certainly know the verb \word{hablar} ``to speak'', but they will only have seen the less frequent, first-person plural future form \word{hablar{\'a}mos} a few times. Nevertheless, they would have no problem translating the latter.
 In this paper we ask whether current methods for bilingual lexicon induction (BLI) generalize morphologically as humans do. Generalization to rare and novel words is arguably the main point of BLI as a task---most frequent translation pairs are already contained in digital dictionaries. Modern word embeddings encode character-level knowledge \cite{bojanowski2017enriching}, which should---in principle---enable the models to learn this behaviour; but morphological generalization has never been directly tested.\looseness=-1

\begin{figure}
\hspace*{-0.63cm}  
  \centering
  \includegraphics[width=1.1\linewidth]{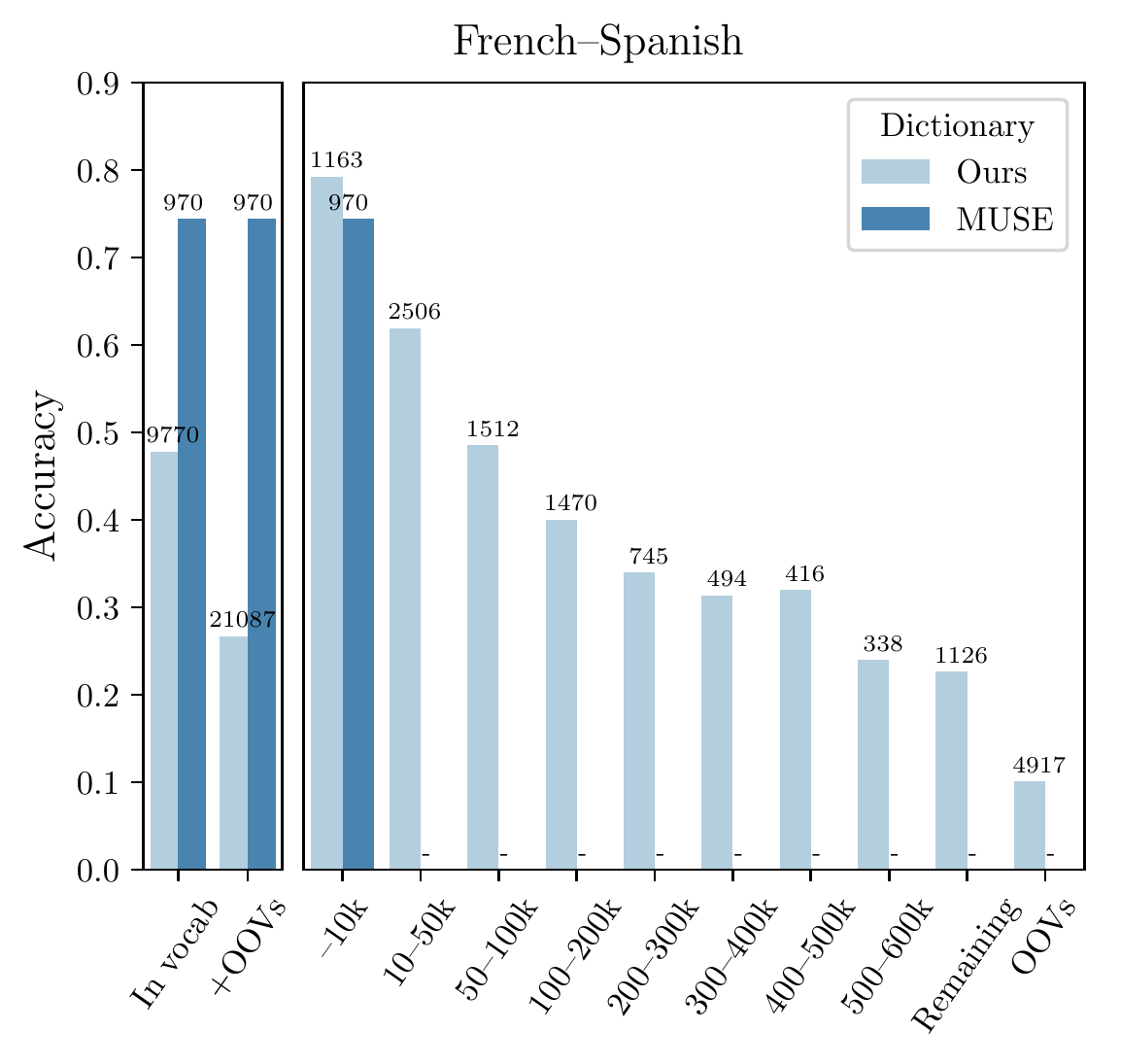}
   \caption{The relation between the BLI performance and the frequency of source words in the test dictionary. The graph presents results for the model of \newcite{ruder2018discriminative} evaluated on both the MUSE dictionary \cite{conneau2018word} and our morphologically complete dictionary, which contains many rare morphological variants of words. The numbers above the bars correspond to the number of translated source words (a hyphen represents an empty dictionary).} % of 
   \label{fig:first-page}
     \vspace*{-0.4cm}  
\end{figure}

Most existing dictionaries used for BLI evaluation do not account for the full spectrum of linguistic properties of language. Specifically, as we demonstrate in \cref{sec:dictionaries}, they omit \emph{most morphological inflections} of even common lexemes. 
To enable a more thorough evaluation we introduce a new resource: 40 \defn{morphologically complete} dictionaries for 5 Slavic and 5 Romance languages, which contain the inflectional paradigm of every word they hold. Much like with a human translator, we expect a BLI model to competently translate full paradigms of lexical items. {Throughout this work we place our focus on genetically-related language pairs. This not only allows us to cleanly map one morphological inflection onto another, but also provides an upper bound for the performance on the generalization task; if the models are not able to generalize for closely related languages they would most certainly be unable to generalize when translating between unrelated languages.}

We use our dictionaries to train and evaluate three of the best performing BLI models \cite{artetxe2016learning,artetxe2017learning,ruder2018discriminative} on all 40 language pairs.  To paint a complete picture of the models' generalization ability we propose a new experimental paradigm in which we independently control for four different variables: the word form's frequency, morphology, the lexeme frequency and the lexeme (a total of 480 experiments). Our comprehensive analysis reveals that BLI models can generalize for frequent morphosyntactic categories, even of infrequent lexemes, but fail to generalize for the more rare categories. This yields a more nuanced picture of the known deficiency of word embeddings to underperform on infrequent words \cite{Gong2018}. Our findings also contradict the strong empirical claims made elsewhere in the literature \cite{artetxe2017learning, conneau2018word, ruder2018discriminative, grave2018unsupervised}, as we observe that performance severely degrades when the evaluation includes rare morphological variants of a word and infrequent lexemes. We picture this general trend in Figure \ref{fig:first-page}, which also highlights the skew of existing dictionaries towards more frequent words.\footnote{For MUSE, we only evaluate on forms that are not present in our training dictionary (970 out of 1500 source words).} As our final contribution, we demonstrate that better encoding of morphology is indeed beneficial: enforcing a simple morphological constraint yields consistent performance improvements for all Romance language pairs and many of the Slavic language pairs.\looseness=-1

\section{Morphological Dictionaries}\label{sec:dictionaries}

\subsection{Existing Dictionaries}  \label{sec:existing_dicts}
Frequent word forms can often be found in human-curated dictionaries. Thus, the practical purpose of training a BLI model should be to create translations of new and less common forms, not present in the existing resources. In spite of this, most ground truth lexica used for BLI evaluation contain mainly frequent word forms. 
Many available resources are restricted to the top 200k most frequent words; this applies to the English--Italian dictionary of  \citet{dinu2015improving},  the English--German and English--Finnish dictionaries of \citet{artetxe2017learning}, and \citet{artetxe2018generalizing}'s English--Spanish resource. The dictionaries of \citet{irvine2017comprehensive} contain only the top most frequent 10k words for each language. \citet{zhang2017adversarial} extracted their Spanish--English and Italian--English lexica from Open Multilingual WordNet \cite{bond2012survey, miller1998wordnet}, a resource which only yields high frequency, lemma level mappings. Another example is the recent MUSE dataset \cite{conneau2018word}, which was generated using an ``internal translation tool'', and in which the majority of word pairs consist of forms ranked in the top 10k of the vocabularies of their respective languages.

Another problem associated with existing resources is `semantic leakage' between train and evaluation sets. As we demonstrate in \textsection \ref{sec:muse}, it is common for a single lexeme to appear in both train and test dictionary---in the form of different word inflections. This circumstance is undesirable in evaluation settings as it can lead to performance overstatements---a model can `memorize' the corresponding target lemma, which ultimately reduces the translation task to a much easier task of finding the most appropriate inflection. Finally, most of the available BLI resources include English in each language pair and, given how morphologically impoverished English is, those resources are unsuitable for analysis of morphological generalization.

\begin{table}
\footnotesize
\centering
\begin{adjustbox}{width=\columnwidth}
\begin{tabular}{l l l l l}
\toprule
\makecell{Source\\Word} & \makecell{Target\\Word}  & 	\makecell{Source\\Lemma}  &	 \makecell{Target\\Lemma} & Tag\\
\midrule
morza	 & mo\v{r}e & 	morze &	mo\v{r}e	&  \T{N;NOM;PL}\\
morzu	&mo\v{r}i & 	morze	 & mo\v{r}e & 	\T{N;DAT;SG} \\
morze	 & mo\v{r}e & 	morze & 	mo\v{r}e	& \T{N;NOM;SG} \\
morzami	 & mo\v{r}i &	morze &	mo\v{r}e	& \T{N;INS;PL} \\
m\'{o}rz	& mo\v{r}í	& morze &	mo\v{r}e	&  \T{N;GEN;PL} \\
morzu &	mo\v{r}i	& morze	& mo\v{r}e	&  \T{N;ESS;SG} \\
morzom & 	mo\v{r}\'{i}m & morze &	 mo\v{r}e & \T{N;DAT;PL} \\
\bottomrule
\end{tabular}
\end{adjustbox}
\caption{An example extract from our morphologically complete Polish--Czech dictionary.}
\label{t:dictexample}
\end{table}

\begin{table*}
\renewcommand\arraystretch{1.2}
\footnotesize
\centering
\begin{tabular}{l  l l l l  | l  l l l l }
\toprule
\textbf{Slavic} &  \lang{Czech}  &  \lang{Russian} &  \lang{Slovak} &   \lang{Ukrainian}  &  
  \textbf{Romance}  & \lang{Spanish} &  \lang{Catalan} &  \lang{Portuguese} &  \lang{Italian}  \\
\midrule
 \lang{Polish}  & 53,353 & 128,638 & 14,517 & 12,361 & 	\lang{French} & 686,139 & 381,825 & 486,575  & 705,800\\
 \lang{Czech}  & - & 65,123 & 10,817 & 8,194 		&\lang{Spanish}  & - & 343,780 & 476,543 & 619,174 \\
 \lang{Russian}  & & - & 128,638 & 10,554 &		\lang{Catalan}  & & - & 261,016 & 351,609  \\
 \lang{Slovak}  & & & - & 3,434 &    		\lang{Portuguese}  & & & - & 468,945 \\
\bottomrule
\end{tabular}
\caption{The sizes of our morphologically complete dictionaries for Slavic and Romance language families. We present the sizes for 20 \emph{base} dictionaries. We further split those to obtain 40 train, development and test dictionaries---one for each mapping direction to ensure the correct source language lemma separation.}
\label{t:dictsize}

\end{table*}
 \subsection{Our Dictionaries}
To address the shortcomings of the existing evaluation, we built 40 new \defn{morphologically complete} dictionaries, which contain most of the inflectional paradigm of every word they contain. This enables a more thorough evaluation and makes the task much more challenging than traditional evaluation sets. In contrast to the existing resources our dictionaries consist of many rare forms, some of which are out-of-vocabulary for large-scale word embeddings such as \model{fastText}. Notably, this makes them the only resource of this kind that enables evaluating \emph{open-vocabulary} BLI.

We focus on pairs of genetically-related languages for which we can cleanly map one morphological inflection onto another.\footnote{One may translate \word{talked}, the past tense of \word{talk}, into many different Spanish forms, but the Portuguese \word{falavam} has, arguably, only one best Spanish translation: \word{hablaban}.} We selected 5 languages from the Slavic family: Polish, Czech, Russian, Slovak and Ukrainian, and 5 Romance languages: French, Spanish, Italian, Portuguese and Catalan. Table \ref{t:dictexample} presents an example extract from our resource; every source--target pair is followed by their corresponding lemmata and a shared tag.

We generated our dictionaries automatically based on openly available resources: {Open Multilingual WordNet} \cite{bond2012survey} and {Extended Open Multilingual WordNet}\footnote{We used the union of the extended and the original version of the multilingual WordNet since the latter included entries that were not present in the extended version.}  \cite{bond2013linking}, both of which are collections of lexical databases which group words into sets of synonyms (synsets), and {UniMorph}\footnote{\url{https://unimorph.github.io/}} \cite{kirov2016very}---a resource comprised of inflectional word paradigms for 107 languages, extracted from Wiktionary\footnote{Wiktionary (\url{https://en.wiktionary.org/wiki/Wiktionary:Main_Page}) is a large-scale, free content multilingual dictionary. (CC BY-SA 3.0); \url{https://creativecommons.org/licenses/by-sa/3.0/}.} and annotated according to the UniMorph schema \cite{sylak2016composition}. For each language pair $(L1, L2)$ we first generated lemma translation pairs by mapping all $L1$ lemmata to all $L2$ lemmata for each synset that appeared in both $L1$ and $L2$ WordNets.\footnote{Note that with this many-to-many mapping we allow for many translations of a single word.} We then filtered out the pairs which contained lemmata not present in UniMorph and generated inflected entries from the remaining pairs: one entry for each tag that appears in the UniMorph paradigms of both lemmata.\footnote{As UniMorph annotations are not consistent across different languages (e.g. in Czech and Polish resources verbs are marked for animacy, while for other Slavic resources they lack such markings) we performed minor tag processing in order to make the tags more compatible. We also discovered that for some languages, the UniMorph resource was either incomplete (resources for Romance languages do not contain adjectives or nouns) or incorrect (Czech verb inflections), in which cases we personally scraped lemma inflections directly from Wiktionary.}
The sizes of dictionaries vary across different language pairs and so does the POS distribution. In particular, while Slavic dictionaries are dominated by nouns and adjectives, verbs constitute the majority of pairs in Romance dictionaries. We report the sizes of the dictionaries in Table \ref{t:dictsize}. In order to prevent semantic leakage, discussed in \textsection \ref{sec:existing_dicts}, for each language pair we split the initial dictionary into train, development and test splits so that each sub-dictionary has its own, independent set of lemmata. In our split, the train dictionary contains 60\% of all lemmata, while the development and test dictionaries each have 20\% of the lemmata.

 \subsection{Comparison with MUSE} \label{sec:muse}
In this section we briefly outline important differences between our resource and the MUSE dictionaries \cite{conneau2018word} for Portuguese, Italian, Spanish, and French (12 dictionaries in total). We focus on MUSE as it is one of the few openly available resources that covers genetically-related language pairs.\looseness=-1 
\paragraph{Word Frequency} 
The first and most prominent difference lies in the skew towards frequent word forms in MUSE evaluation. While our test dictionaries contain a representative sample of forms in lower frequency bins, the majority of forms present in MUSE are ranked in the top 10k  in their respective language vocabularies. This is clearly presented in Figure \ref{fig:first-page} for the French--Spanish resource and also holds for the remaining 11 dictionaries.
\paragraph{Morphological Diversity} Another difference lies in the morphological diversity of both dictionaries. The average proportion of paradigm covered for lemmata present in MUSE test dictionaries is 53\% for nouns, 37\% for adjectives and only 3\% for verbs. We generally observe that for most lemmata the dictionaries contain \emph{only one} inflection.  In contrast, {for our test dictionaries we get 97\% coverage for nouns, 98\% for adjectives and 67\% for verbs.} Note that we do not get 100\% coverage as we are limited by the compatibility of source language and target language UniMorph resources.
\paragraph{Train--test Paradigm Leakage} Finally, we carefully analyze the magnitude of the train--test paradigm leakage. We found that, on average 20\% (299 out of 1500) of source words in MUSE test dictionaries share their lemma with a word in the corresponding train dictionary. E.g. the French--Spanish test set includes the form \word{perdent}---a third-person plural present indicative of \word{perdre} (to lose) which is present in the train set. 
Note that the splits we provide for our dictionaries do not suffer from any leakage as we ensure that each dictionary contains the full paradigm of every lemma.

\begin{figure*}
\begin{subfigure}{.5\textwidth}
\hspace*{-5mm}  
  \centering
  \includegraphics[width=1.05\linewidth]{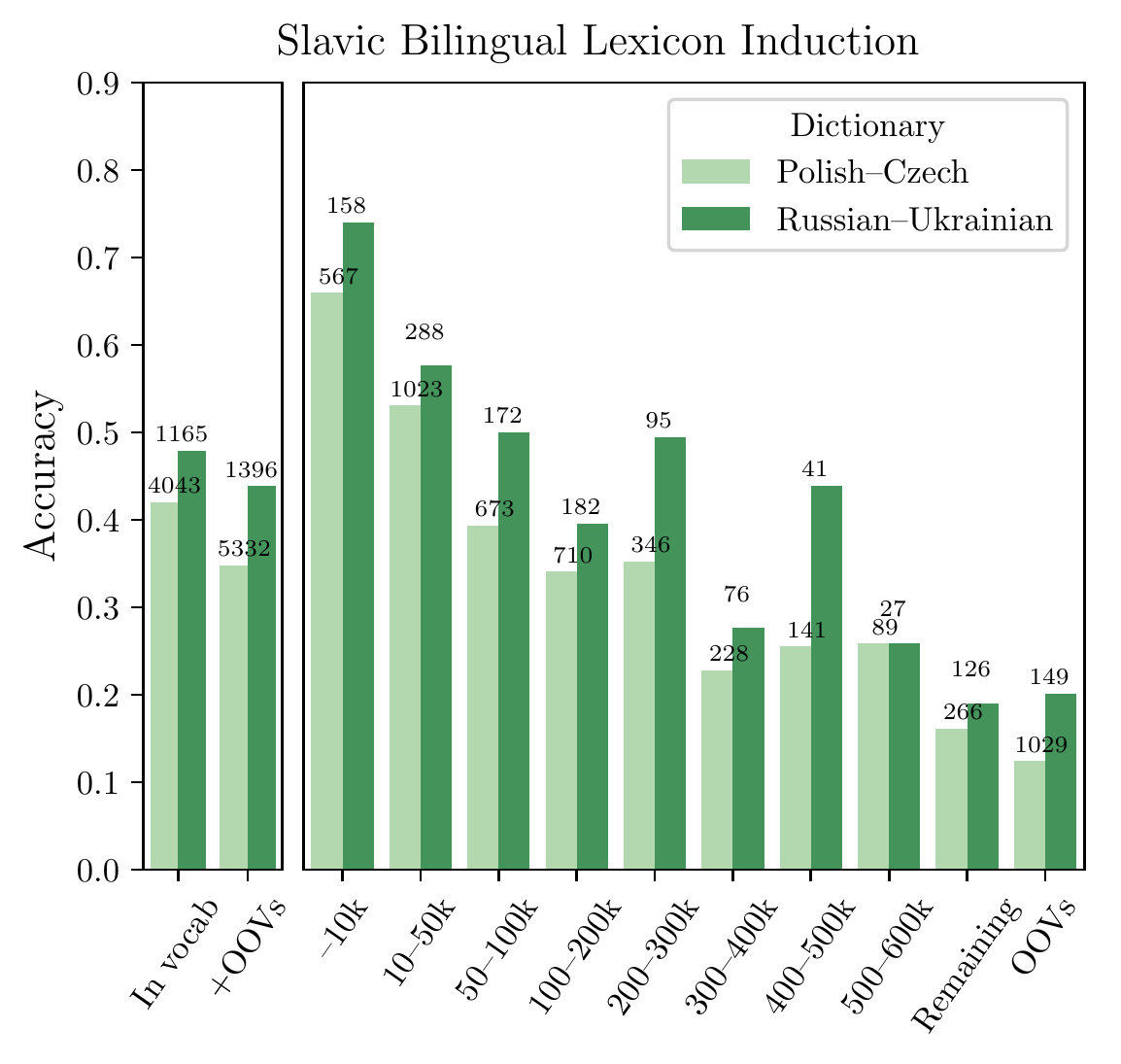}
\end{subfigure}%
\begin{subfigure}{.5\textwidth}
\hspace*{-0.25cm}  
  \centering
  \includegraphics[width=1.05\linewidth]{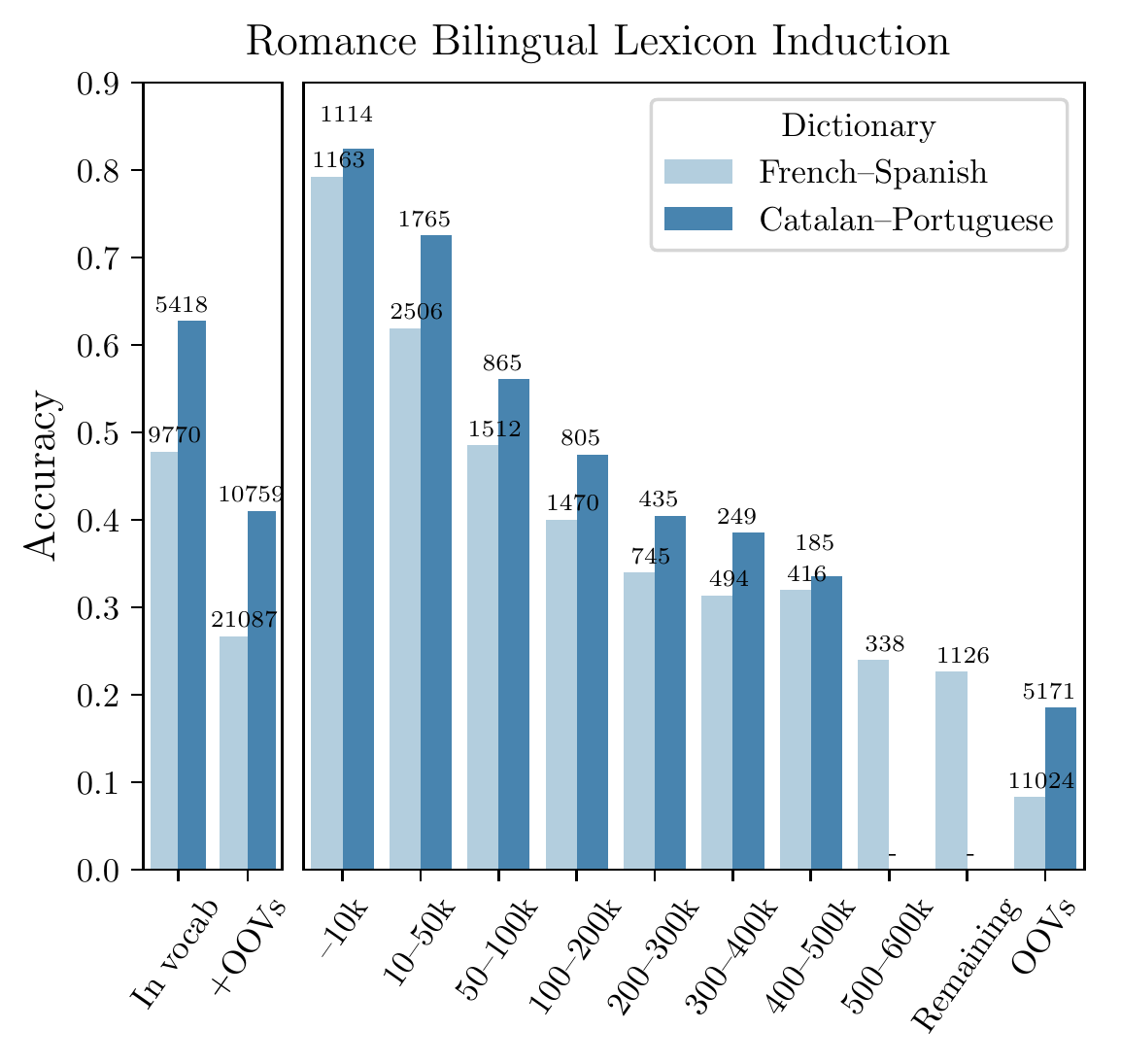}
\end{subfigure}
\caption{The relation between performance and the frequency of source words in the test dictionary for four example language pairs on the standard BLI task. The numbers above the bars correspond to the dictionary sizes.}
\label{fig:frequency-bars}
\end{figure*}

\section{Bilingual Lexicon Induction} \label{sec:exp1}
The task of bilingual lexicon induction is well established in the community \cite{Vulic2013a,Gouws2015} and is the current standard choice for evaluation of cross-lingual word embedding models. Given a list of $N$ source language word forms $x_1, \ldots, x_N$, the goal is to determine the most appropriate translation $t_i$, for each query form $x_i$. In the context of cross-lingual embeddings, this is commonly accomplished by finding a target language word that is most similar to $x_i$ in the shared semantic space, where words' similarity is usually computed using a cosine between their embeddings. 
The resulting set of $(x_i, t_i)$ pairs is then compared to the gold standard and evaluated using the precision at $k$ (P@$k$) metric, where $k$ is typically set to 1, 5 or 10.\footnote{Precision at $k$ represents how many times the correct translation of a source word is returned as one of its $k$ nearest neighbours in the target language.} Throughout our evaluation we use P@1, which is equivalent to accuracy.

In our work, we focus on the supervised and semi-supervised settings in which the goal is to automatically generate a dictionary given only monolingual word embeddings and some initial, seed translations. 
For our experiments we selected the models of \citet{artetxe2016learning}, \citet{artetxe2017learning} and \citet{ruder2018discriminative}---three of the best performing BLI models, which induce a shared cross-lingual embedding space by learning an orthogonal transformation from one monolingual space to another (model descriptions are given in the supplementary material).\footnote{We also used our dictionaries to train and test the more recent model of \citet{artetxe2018robust} on a handful of languages and observed the same general trends.} In particular, the last two employ a self-learning method in which they alternate between a mapping step and a word alignment (dictionary induction) step in an iterative manner. As we observed the same general trends across all models, in the body of the paper we only report the results for the best performing model of \citet{ruder2018discriminative}. We present the complete set of results in the supplementary material.

\paragraph{Experimental Setup} We trained and evaluated all models using the Wikipedia \model{fastText} embeddings \cite{grave2018learning}. Following the existing work, for training we only used the most frequent 200k words in both source and target vocabularies. To allow for evaluation on less frequent words, in all our experiments the models search through the whole target embedding matrix at evaluation (not just the top 200k words, as is common in the literature). This makes the task more challenging, but also gives a more accurate picture of performance. 
To enable evaluation on the unseen word forms we generated a \model{fastText} embedding for every out-of-vocabulary (OOV) inflection of every word in WordNet that also appears in UniMorph. We built those embeddings by summing the vectors of all $n$-grams that constitute an OOV form.\footnote{We did this within the \model{fastText} framework, using the trained $.bin$ models for each of our 10 languages.}
In the OOV evaluation we append the resulting vectors to the original embedding matrices.\looseness=-1

\section{Morphological Generalization} \label{sec:morph_gen}
We propose a novel \defn{quadripartite analysis} of the BLI models, in which we independently control for four different variables: (i) word form frequency, (ii) morphology, (iii) lexeme frequency and (iv) lexeme. We provide detailed descriptions for each of those conditions in the following sections. For each condition, we analyzed all 40 language pairs for each of our selected models---a total of 480 experiments. In the body of the paper we only present \emph{a small representative subset} of our results.

\subsection{Controlling for Word Frequency} \label{sec:freq_control}
For highly inflectional languages, many of the infrequent types are rare forms of otherwise common lexemes and, given the morphological regularity of less frequent forms, a model that generalizes well should be able to translate those capably. Thus, to gain insight into the models' generalization ability we first examine the relation between their performance and the frequency of words in the test set.

We split each test dictionary into 9 frequency bins, based on the relative frequencies of words in the original training corpus for the word embeddings (Wikipedia in the case of \model{fastText}). More specifically, a pair appears in a frequency bin if its source word belongs to that bin, according to its rank in the respective vocabulary.  
We also considered unseen words that appear in the test portion of our dictionaries, but do not occur in the training corpus for the embeddings. This is a fair experimental setting since most of those OOV words are associated with known lemmata. Note that it bears a resemblance to the classic Wug Test \cite{berko1958child} in which a child is introduced to a single instance of a fictitious object---`a wug'---and is asked to name two instances of the same object---`wugs'. However, in contrast to the original setup, we are interested in making sure the unseen inflection of a known lexeme is properly translated.

Figure \ref{fig:frequency-bars} presents the results on the BLI task for four example language pairs: two from the Slavic and two from the Romance language family. The left-hand side of the plots shows the performance for the full dictionaries (with and without OOVs), while the right-hand side demonstrates how the performance changes as the words in the evaluation set become less frequent. The general trend we observe across all language pairs is an acute drop in accuracy for infrequent word forms---e.g. for Catalan--Portuguese the performance falls {from 83\% for pairs containing only the top 10k most frequent words to 40\% for pairs, which contain source words ranked between 200k and 300k}.

\begin{table*}[h]
\footnotesize
\begin{tabular}{l | ll | rrrrrrrrrr}
\toprule
\multirow{2}{*}{Tag}            & \multicolumn{2}{c |}{Accuracy} &  \multicolumn{10}{c }{Distribution Across Frequency Bins} \\
& In vocab &  All  &   10k &   50k &   100k &   200k &   300k &   400k &   500k &   600k &   Tail &   OOVs \\
\midrule
 \multicolumn{13}{c}{French--Spanish} \\
\midrule
\T{N;SG}           &       56.7 &              55.1 & 45\%  & 33\%  & 10\%   & 6\%    & 1\%    & 1\%    & 0\%    & 0\%    & 1\%    & 2\%    \\
\T{N;PL}          &       53.6 &              49.9 & 31\%  & 29\%  & 14\%   & 9\%    & 4\%    & 3\%    & 2\%    & 1\%    & 3\%    & 4\%    \\
\T{ADJ;MASC;SG}    &       62.7 &              60.9 & 29\%  & 40\%  & 16\%   & 7\%    & 3\%    & 1\%    & 1\%    & 0\%    & 1\%    & 2\%    \\
\T{NFIN;V}         &       50.6 &              49.0 & 35\%  & 37\%  & 13\%   & 7\%    & 2\%    & 1\%    & 1\%    & 1\%    & 2\%    & 2\%    \\
\T{2;IMP;PL;POS;V} &        1.5 &               2.0 & 3\%   & 4\%   & 16\%   & 14\%   & 12\%   & 4\%    & 5\%    & 3\%    & 12\%   & 25\%   \\
\T{3;FUT;PL;V}     &       34.7 &              23.3 & 0\%   & 4\%   & 8\%    & 14\%   & 9\%    & 8\%    & 8\%    & 5\%    & 16\%   & 27\%   \\
\T{3;COND;SG;V}    &       41.0 &              30.6 & 0\%   & 5\%   & 14\%   & 14\%   & 9\%    & 9\%    & 3\%    & 6\%    & 15\%   & 23\%   \\
\midrule

 \multicolumn{13}{c}{Polish--Czech} \\
\midrule
\T{INS;N;PL}        &       29.5 &              23.8  & 1\%   & 17\%  & 16\%   & 21\%   & 10\%   & 7\%    & 7\%    & 2\%    & 9\%    & 10\%   \\
\T{DAT;N;SG}        &       26.1 &              18.7 & 13\%  & 15\%  & 10\%   & 12\%   & 7\%    & 4\%    & 3\%    & 3\%    & 8\%    & 24\%   \\
\T{ACC;ADJ;MASC;SG} &       47.8 &              47.8 & 10\%  & 28\%  & 27\%   & 16\%   & 11\%   & 4\%    & 1\%    & 1\%    & 1\%    & 0\%    \\
\T{ACC;ADJ;FEM;SG}  &       47.0 &              46.4 & 4\%   & 29\%  & 27\%   & 25\%   & 6\%    & 4\%    & 3\%    & 1\%    & 0\%    & 0\%    \\
\T{NFIN;V}          &       47.4 &              45.0    & 13\%  & 42\%  & 18\%   & 15\%   & 4\%    & 4\%    & 0\%    & 0\%    & 0\%    & 5\%    \\
\T{3;MASC;PL;PST;V} &       59.4 &              51.4 & 2\%   & 23\%  & 16\%   & 21\%   & 9\%    & 12\%   & 5\%    & 5\%    & 0\%    & 7\%    \\
\T{2;IMP;PL;V}      &       11.1 &               8.1 & 0\%   & 0\%   & 0\%    & 14\%   & 0\%    & 7\%    & 14\%   & 7\%    & 3\%    & 55\%   \\

\bottomrule

\end{tabular}
\caption{ BLI results for word pairs that have a specific morphosyntactic category (left) and a distribution of those forms across different frequency bins (right).}
\label{tab:morph_dist}
\end{table*}

\subsection{Controlling for {Morphology}} \label{sec:morph_control}

From the results of the previous section, it is not clear whether the models perform badly on inflections of generally \emph{infrequent lemmata} or whether they fail on \emph{infrequent morphosyntactic categories}, independently of the lexeme frequency. Indeed, the frequency of different morphosyntactic categories is far from uniform. To shed more light on the underlying cause of the performance drop in \cref{sec:freq_control}, we first analyze the differences in the models' performance as they translate forms belonging to different categories and, next, look at the distribution of these categories across the frequency bins.

In Table \ref{tab:morph_dist} we present our findings for a representative sample of morphosyntactic categories for one Slavic and one Romance language pair (we present the results for all models and all language pairs in the supplementary material).\footnote{We selected the morphosyntactic categories independently for each language family based on how well the models translate words of such morphology--`the good, the bad and the ugly'.} It illustrates the great variability across different paradigm slots---both in terms of their frequency and the difficulty of their translation.

As expected, the performance is best for the slots belonging to the highest frequency bins and forms residing in the rarer slots prove to be more challenging. For example, for French--Spanish the performance on \T{2;IMP;PL;POS;V}, \T{3;FUT;PL;V} and \T{3;COND;SG;V}\footnote{2nd person imperative plural form of a verb, 3rd person future plural form of a verb and 2nd person conditional singular form of a verb.} is notably lower than that for the remaining categories. For both language pairs, the accuracy for the second-person plural present imperative (\T{2;IMP;PL;V}) is particularly low: {1.5\% accuracy for French--Spanish and 11.1\% for Polish--Czech} in the in-vocabulary setting. Note that it is unsurprising for an imperative form, expressing an order or command, to be infrequent in the Wikipedia corpora (the resource our monolingual embeddings were trained on). {The complex distribution of the French \T{2;IMP;PL;POS;V} across the frequency bins is likely due to syncretism---the \T{2;IMP;PL;POS;V} paradigm slot shares a form with 2nd person present plural slot, \T{2;PL;PRS;V}.  Our hypothesis is that syncretism may have an effect on the quality of the monolingual embeddings. To our knowledge, the effect of syncretism on embeddings has not yet been systematically investigated.
}

\begin{figure*}[h]
\begin{subfigure}{.5\textwidth}
\hspace*{-0.6cm}  
  \centering
  \includegraphics[width=1.06\linewidth]{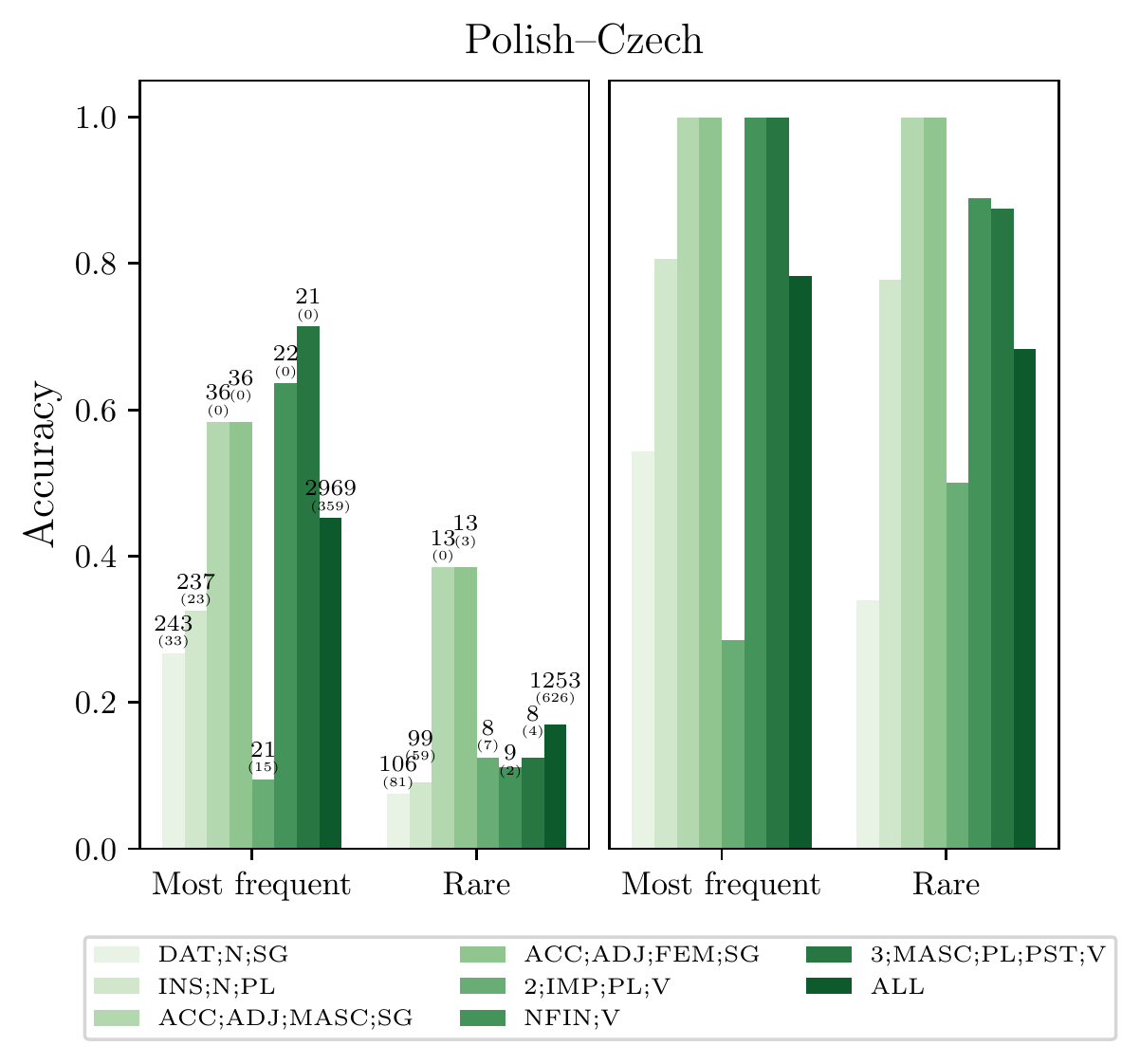}
\end{subfigure}%
\begin{subfigure}{.5\textwidth}
\hspace*{-0.26cm}  
  \centering
 \includegraphics[width=1.04\linewidth]{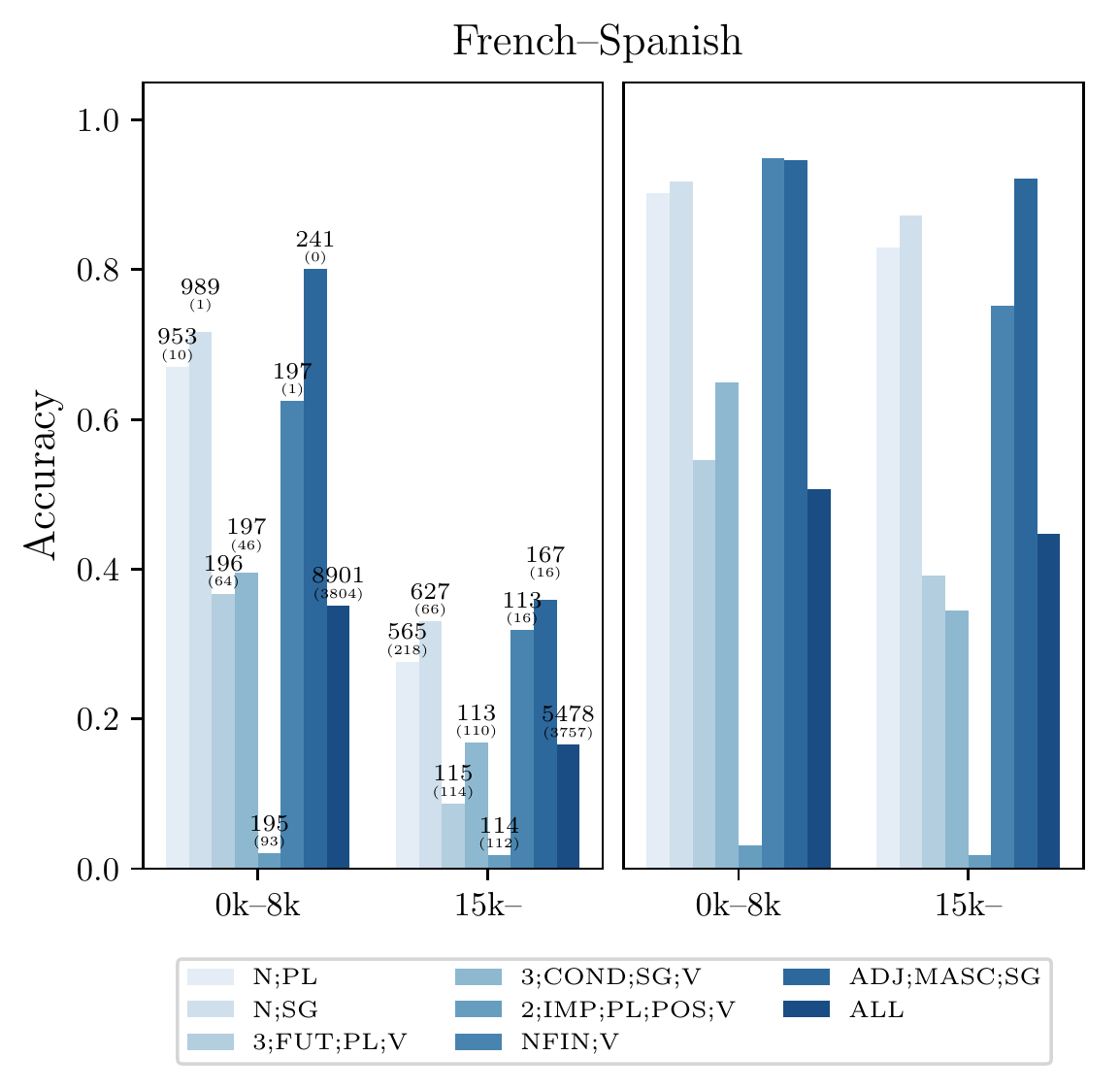}  
\end{subfigure}

\caption{The performance on the standard BLI task (left side of the graphs) and the controlled for lexeme BLI (right side) for words pairs belonging to the most frequent paradigms and the infrequent paradigms. The numbers above the bars are dictionary sizes and the number of out-of-vocabulary forms in each dictionary (bracketed).}
\label{fig:morph-steps}
%%\vspace{-4mm}
\end{figure*}

\subsection{Controlling for Lexeme Frequency} \label{sec:par_control}
To get an even more complete picture, we inspect how the performance on translating inflections of \emph{common lemmata} differs to translating forms coming from \emph{less frequent paradigms} by controlling for the frequency of the lexeme. 
We separated our dictionaries into two bins based on the relative frequency of the source lexeme. We approximated frequency of the lexemes by using ranks of their most common inflections: our first bin contained lexemes whose most common inflection is ranked in the top 20k forms in its respective vocabulary, while the second bin consisted of lexemes with most common inflection ranked lower than 60k.\footnote{Future work could operate on actual word counts rather than relative frequencies of words.}
We present the results for the same morphosyntactic categories as in \textsection \ref{sec:morph_control} on the left side of the graphs in Figure \ref{fig:morph-steps}. As anticipated, in the case of less frequent lexemes the performance is generally worse than for frequent ones. However, perhaps more surprisingly, we discover that some morphosyntactic categories prove to be problematic even for the most frequent lexemes. Some examples include the previously mentioned imperative verb form or, for Slavic languages, singular dative nouns (\T{DAT;N;SG}).

\begin{table*}[htbp]
\centering
\small
\renewcommand\arraystretch{1.2}
\begin{tabular}{ l ||  c  c | c c || c c | c c || c c}
\toprule 
 & \multicolumn{4}{c ||}{Normal} & \multicolumn{4}{c ||}{Lexeme} & \multicolumn{2}{c }{Dictionary Sizes}  \\ 
 & \multicolumn{2}{c}{In vocab} & \multicolumn{2}{ c || }{+OOVs} &  \multicolumn{2}{ c}{In vocab} & \multicolumn{2}{c ||}{+OOVs} & \multirow{ 2}{*}{In vocab} & \multirow{ 2}{*}{+OOVs}\\ 
 
\multirow{3}{*}{}Constraint & \xmark & \cmark & \xmark & \cmark & \xmark & \cmark & \xmark & \cmark \\ 
 \hline
 Ukrainian--Russian  & \B{68.4} & 61.1 & \B{63.7} & 56.1 & \B{89.9} & 89.1 & \B{88.6} & 87.6 & 786 & 933 \\
 Russian--Slovak     & \B{25.7} & 21.1 & \B{20.9} & 17.0 & \B{79.3} & 76.8 & \B{76.0} & 74.2 & 1610 & 2150 \\
 Polish--Czech       & 42.0 & \B{44.4} & 34.8 & \B{36.7} & 80.6 & \B{81.1} & 75.3 & \B{75.9} & 4043 & 5332 \\
 Russian--Polish     & 39.8 & \B{41.2} & 34.8 & \B{36.1} & 80.8 & \B{82.6} & 77.7 & \B{80.2} & 9183 & 11697 \\
 \hline
 Catalan--Portuguese & 62.8 & \B{64.2} & 41.1 & \B{42.4} & 83.1 & \B{84.3} & 57.7 & \B{59.0} & 5418 & 10759 \\
 French--Spanish     & 47.8 & \B{50.2} & 26.7 & \B{28.9} & 78.0 & \B{81.4} & 47.9 & \B{52.2} & 9770 & 21087 \\
 Portuguese--Spanish & 60.2 & \B{61.1} & 36.8 & \B{37.6} & 84.7 & \B{85.4} & 57.1 & \B{58.2} & 9275 &  22638\\
 Italian--Spanish    & 42.7 & \B{43.8} & 21.1 & \B{22.1} & 76.4 & \B{77.6} & 47.6 & \B{49.6} & 11685 &  30686\\
 \hline
 Polish--Spanish     & \B{36.1}  & 32.1 & \B{28.0} & 25.0 & \B{78.1} & 77.7 & \B{68.6} & 68.4 & 8964 & 12759\\
 Spanish--Polish     & 28.1  & \B{30.9} & 21.0  & \B{23.2} & 81.2 & \B{82.0} & 64.2 & \B{65.8} & 4270 & 6095 \\

\end{tabular}
\caption{The results on the standard BLI task and BLI controlled for lexeme for the original \citet{ruder2018discriminative}'s model (\xmark) and the same model trained with a morphological constraint (\cmark) (discussed in \S\ref{sec:exp3}).}
\label{t:all_res}
\end{table*}

\subsection{Controlling for Lexeme}\label{sec:exp2}

We are, in principle, interested in the ability of the models to generalize \emph{morphologically}. In the preceding sections we focused on the standard BLI evaluation, which given our objective is somewhat unfair to the models---they are additionally punished for not capturing lexical semantics. To gain more direct insight into the models' generalization abilities we develop a novel experiment in which the lexeme is controlled for. At test time, the BLI model is given a set of candidate translations, all of which belong to the same paradigm, and is asked to select the most suitable form. Note that the model only requires morphological knowledge to successfully complete the task---no lexical semantics is required. When mapping between closely related languages this task is particularly straightforward, and especially so in the case of \model{fastText} where a single $n$-gram, e.g. the suffix \word{-ing} in English as in the noun \word{running}, can be highly indicative of the inflectional morphology of the word.

We present results on 8 representative language pairs in Table \ref{t:all_res} (column Lexeme). We report the accuracy on the in-vocabulary pairs as well as all the pairs in the dictionary, including OOVs. As expected, compared to standard BLI this task is much easier for the models---the performance is generally high. For Slavic languages numbers remain high even in the open-vocabulary setup, which suggests that the models can generalize morphologically. On the other hand, for Romance languages we observe a visible drop in performance. 
We hypothesize that this difference is due to the large quantities of verbs in Romance dictionaries; in both Slavic and Romance languages verbs have substantial paradigms, often of more than 60 forms, which makes identifying the correct form more difficult. In contrast, most words in our Slavic dictionaries are nouns and adjectives with much smaller paradigms.

Following our analysis in \cref{sec:par_control}, we also examine how the performance on this new task differs for less and more frequent paradigms, as well as across different morphosyntactic categories. Here, we exhibit an unexpected result, which we present in the two right-hand side graphs of Figure \ref{fig:morph-steps}: the state-of-the-art BLI models \emph{do} generalize morphologically for frequent slots, but \emph{do not} generalize for infrequent slots. 
{For instance, for the Polish--Czech pair, the models achieve 100\% accuracy on identifying the correct inflection when this inflection is {\T{ACC;ADJ;MASC;SG}, \T{ACC;ADJ;FEM;SG}, \T{3;MASC;PL;PST;V} or \T{NFIN;V}}\footnote{Masculine accusative singular form of an adjective, feminine accusative singular form of an adjective, 3rd person masculine plural past form of a verb and a verbal infinitive.} for frequent and, for the first two categories, also  the \emph{infrequent} lexemes; all of which are common morphosyntactic categories (see Table \ref{tab:morph_dist}). The results from Figure \ref{fig:morph-steps} also demonstrate that the worst performing forms for the French--Spanish language pair are indeed the infrequent verbal inflections.}

\subsection{Experiments on an Unrelated Language Pair}\label{sec:unrelated}
So far, in our evaluation we have focused on pairs of genetically-related languages, which provided an upper bound for morphological generalization in BLI.
But our experimental paradigm is not limited to related language pairs. We demonstrate this by experimenting on two example pairs of one Slavic and one Romance language: Polish--Spanish and Spanish--Polish. To construct the dictionaries we followed the procedure discussed in \textsection \ref{sec:dictionaries}, but matched the tags based only on the features exhibited in both languages (e.g. Polish \T{DAT;N;SG} can be mapped to \T{N;SG} in Spanish, as Spanish nouns are not declined for case). Note that mapping between morphosyntactic categories of two unrelated languages is a challenging task \cite{haspelmath2010comparative}, but we did our best to address the issues specific to translation between Polish and Spanish. E.g. we ensured that Spanish imperfective/perfective verb forms can only be translated to Polish forms of imperfective/perfective verbs.

\begin{figure}
\hspace*{-0.45cm}  
  \centering
  \includegraphics[width=1.08\linewidth]{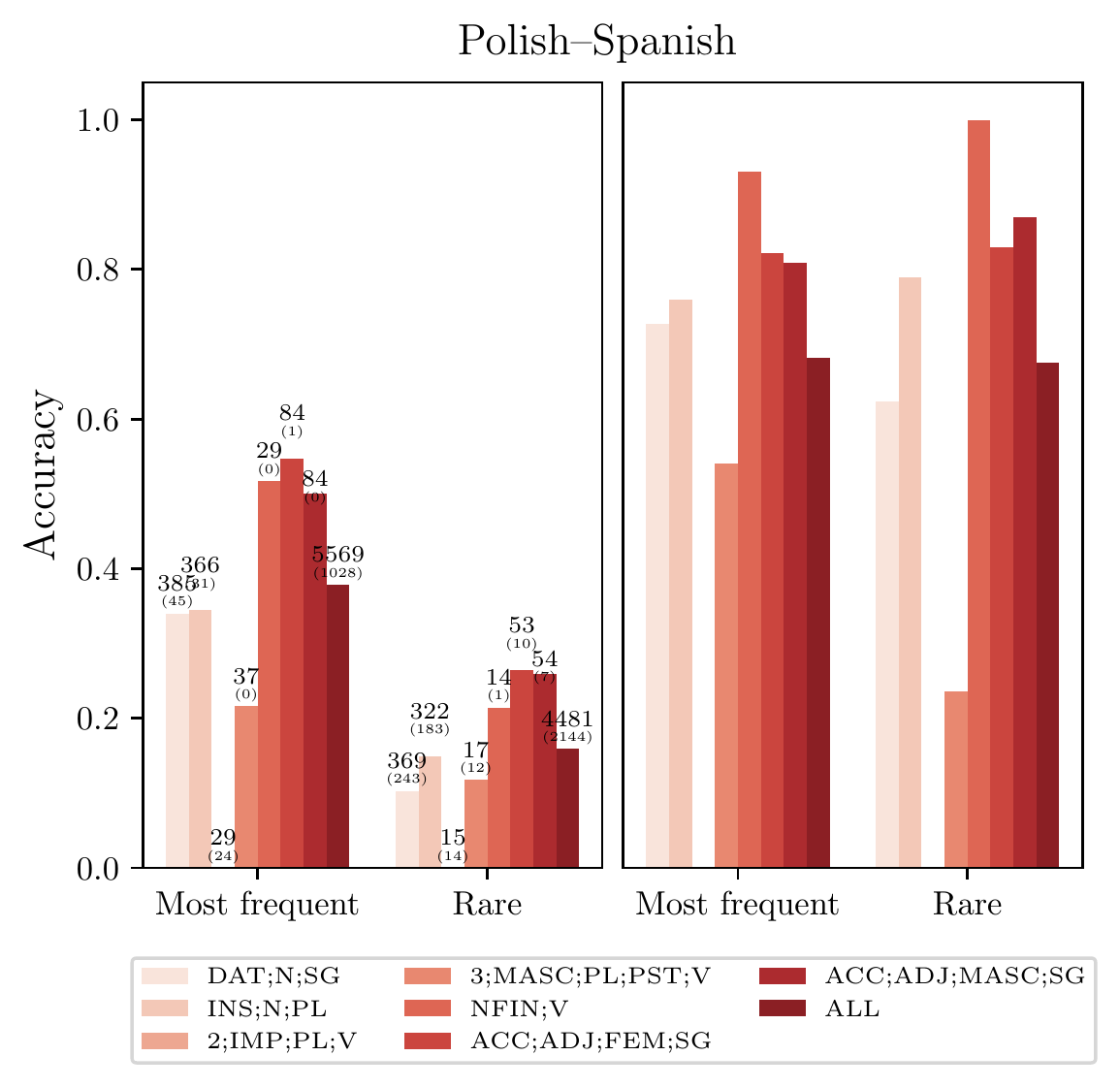}
\caption{The results of the experiments on a pair of unrelated languages---Polish and Spanish---on the standard BLI task (left side) and the controlled for lexeme BLI (right side) for word pairs belonging to the most frequent and the infrequent paradigms.}
\label{fig:morph-unrel}
\vspace{-0.2cm}
\end{figure}

The results of our experiments are presented in the last two rows of Table \ref{t:all_res} and, for Polish--Spanish, also in Figure \ref{fig:morph-unrel}. As expected, the BLI results on unrelated languages are generally, but not uniformly, worse than those on related language pairs. { The accuracy for Spanish--Polish is particularly low, at 28\% (for in vocabulary pairs).} { We see large variation in performance across morphosyntactic categories and more and less frequent lexemes, similar to that observed for related language pairs. In particular, we observe that \T{2;IMP;PL;V}---the category difficult for Polish--Czech BLI is also among the most challenging for Polish--Spanish. However, one of the highest performing categories for Polish--Czech, \T{3;MASC;PL;PST;V}, yields much worse accuracy for Polish--Spanish.} 

\subsection{Adding a Morphological Constraint}\label{sec:exp3}

In our final experiment we demonstrate that improving morphological generalization has the potential to improve BLI results. We show that enforcing a simple, hard morphological constraint \emph{at training time} can lead to performance improvements at test time---both on the standard BLI task and the controlled for lexeme BLI. We adapt the self-learning models of \citet{artetxe2017learning} and \citet{ruder2018discriminative} so that at each iteration they can align two words only if they share the same morphosyntactic category. Note that this limits the training data only to word forms present in UniMorph, as those are the only ones for which we have a gold tag.\footnote{We also experimented with more relaxed versions of the constraint, where we only used a subset of the features, but we found that for most languages the constraint worked better with more features.} The results, a subset of which we present in Table \ref{t:all_res}, show that the constraint, despite its simplicity and being trained on less data, leads to performance improvements for \emph{every} Romance language pair and many of the Slavic language pairs. We take this as evidence that properly modelling morphology will have a role to play in BLI. 

\section{Discussion and Conclusion}

We conducted a large-scale evaluation of the generalization ability of the state-of-the-art bilingual lexicon inducers.
To enable our analysis we created 40 morphologically complete dictionaries for 5 Slavic and 5 Romance languages and proposed a novel experimental paradigm in which we independently control for four different variables.

Our study is the first to examine morphological generalization in BLI and it reveals a nuanced picture of the interplay between performance, the word's frequency and morphology. We observe that the performance degrades when models are evaluated on less common words---even for the infrequent forms of common lexemes. Our results from the controlled for lexeme experiments  suggest that models are able to generalize well for more frequent morphosyntactic categories and for part-of-speech with smaller paradigms. However, their ability to generalize decreases as the slots get less frequent and/or the paradigms get larger. Finally, we proposed a simple method to inject morphological knowledge and demonstrated that making models more morphologically aware can lead to general performance improvements.

\bibliography{emnlp-ijcnlp-2019}

\begin{thebibliography}{23}
\expandafter\ifx\csname natexlab\endcsname\relax\def\natexlab#1{#1}\fi

\bibitem[{Artetxe et~al.(2016)Artetxe, Labaka, and
  Agirre}]{artetxe2016learning}
Mikel Artetxe, Gorka Labaka, and Eneko Agirre. 2016.
\newblock Learning principled bilingual mappings of word embeddings while
  preserving monolingual invariance.
\newblock In \emph{Proceedings of the 2016 Conference on Empirical Methods in
  Natural Language Processing}, pages 2289--2294.

\bibitem[{Artetxe et~al.(2017)Artetxe, Labaka, and
  Agirre}]{artetxe2017learning}
Mikel Artetxe, Gorka Labaka, and Eneko Agirre. 2017.
\newblock \href {https://doi.org/10.18653/v1/P17-1042} {Learning bilingual word
  embeddings with (almost) no bilingual data}.
\newblock In \emph{Proceedings of the 55th Annual Meeting of the Association
  for Computational Linguistics (Volume 1: Long Papers)}, pages 451--462.
  Association for Computational Linguistics.

\bibitem[{Artetxe et~al.(2018{\natexlab{a}})Artetxe, Labaka, and
  Agirre}]{artetxe2018generalizing}
Mikel Artetxe, Gorka Labaka, and Eneko Agirre. 2018{\natexlab{a}}.
\newblock Generalizing and improving bilingual word embedding mappings with a
  multi-step framework of linear transformations.
\newblock In \emph{Thirty-Second AAAI Conference on Artificial Intelligence}.

\bibitem[{Artetxe et~al.(2018{\natexlab{b}})Artetxe, Labaka, and
  Agirre}]{artetxe2018robust}
Mikel Artetxe, Gorka Labaka, and Eneko Agirre. 2018{\natexlab{b}}.
\newblock \href {http://aclweb.org/anthology/P18-1073} {A robust self-learning
  method for fully unsupervised cross-lingual mappings of word embeddings}.
\newblock In \emph{Proceedings of the 56th Annual Meeting of the Association
  for Computational Linguistics (Volume 1: Long Papers)}, pages 789--798.
  Association for Computational Linguistics.

\bibitem[{Berko(1958)}]{berko1958child}
Jean Berko. 1958.
\newblock The child's learning of {E}nglish morphology.
\newblock \emph{Word}, 14(2-3):150--177.

\bibitem[{Bojanowski et~al.(2017)Bojanowski, Grave, Joulin, and
  Mikolov}]{bojanowski2017enriching}
Piotr Bojanowski, Edouard Grave, Armand Joulin, and Tomas Mikolov. 2017.
\newblock \href {https://transacl.org/ojs/index.php/tacl/article/view/999}
  {Enriching word vectors with subword information}.
\newblock \emph{Transactions of the Association for Computational Linguistics},
  5:135--146.

\bibitem[{Bond and Foster(2013)}]{bond2013linking}
Francis Bond and Ryan Foster. 2013.
\newblock Linking and extending an open multilingual wordnet.
\newblock In \emph{Proceedings of the 51st Annual Meeting of the Association
  for Computational Linguistics (Volume 1: Long Papers)}, volume~1, pages
  1352--1362.

\bibitem[{Bond and Paik(2012)}]{bond2012survey}
Francis Bond and Kyonghee Paik. 2012.
\newblock A survey of wordnets and their licenses.
\newblock In \emph{Proceedings of the 6th Global WordNet Conference (GWC
  2012)}, page 64–71.

\bibitem[{Conneau et~al.(2018)Conneau, Lample, Ranzato, Denoyer, and
  J{\'{e}}gou}]{conneau2018word}
Alexis Conneau, Guillaume Lample, Marc'Aurelio Ranzato, Ludovic Denoyer, and
  Herv{\'{e}} J{\'{e}}gou. 2018.
\newblock Word translation without parallel data.
\newblock In \emph{Proceedings of ICLR 2018}.

\bibitem[{Dinu et~al.(2015)Dinu, Lazaridou, and Baroni}]{dinu2015improving}
Georgiana Dinu, Angeliki Lazaridou, and Marco Baroni. 2015.
\newblock Improving zero-shot learning by mitigating the hubness problem.
\newblock In \emph{Proceedings of the 3rd International Conference on Learning
  Representations (ICLR 2015), workshop track, Workshop Track}.

\bibitem[{Gong et~al.(2018)Gong, He, Tan, Qin, Wang, and Liu}]{Gong2018}
Chengyue Gong, Di~He, Xu~Tan, Tao Qin, Liwei Wang, and Tie-Yan Liu. 2018.
\newblock \href {http://arxiv.org/abs/arXiv:1809.06858v1} {{FRAGE:
  Frequency-Agnostic Word Representation}}.
\newblock In \emph{Proceedings of NIPS 2018}.

\bibitem[{Gouws et~al.(2015)Gouws, Bengio, and Corrado}]{Gouws2015}
Stephan Gouws, Yoshua Bengio, and Greg Corrado. 2015.
\newblock \href {http://arxiv.org/abs/1410.2455} {{BilBOWA: Fast Bilingual
  Distributed Representations without Word Alignments}}.
\newblock \emph{Proceedings of The 32nd International Conference on Machine
  Learning}, pages 748--756.

\bibitem[{Grave et~al.(2018{\natexlab{a}})Grave, Bojanowski, Gupta, Joulin, and
  Mikolov}]{grave2018learning}
Edouard Grave, Piotr Bojanowski, Prakhar Gupta, Armand Joulin, and Tomas
  Mikolov. 2018{\natexlab{a}}.
\newblock Learning word vectors for 157 languages.
\newblock In \emph{Proceedings of the International Conference on Language
  Resources and Evaluation (LREC 2018)}.

\bibitem[{Grave et~al.(2018{\natexlab{b}})Grave, Joulin, and
  Berthet}]{grave2018unsupervised}
Edouard Grave, Armand Joulin, and Quentin Berthet. 2018{\natexlab{b}}.
\newblock \href {http://arxiv.org/abs/1805.11222} {Unsupervised alignment of
  embeddings with wasserstein procrustes}.
\newblock \emph{CoRR}, abs/1805.11222.

\bibitem[{Haspelmath(2010)}]{haspelmath2010comparative}
Martin Haspelmath. 2010.
\newblock Comparative concepts and descriptive categories in crosslinguistic
  studies.
\newblock \emph{Language}, 86(3):663--687.

\bibitem[{Irvine and Callison-Burch(2017)}]{irvine2017comprehensive}
Ann Irvine and Chris Callison-Burch. 2017.
\newblock \href {https://doi.org/10.1162/COLI\_a\_00284} {A comprehensive
  analysis of bilingual lexicon induction}.
\newblock \emph{Computational Linguistics}, 43(2):273--310.

\bibitem[{Kirov et~al.(2016)Kirov, Sylak-Glassman, Que, and
  Yarowsky}]{kirov2016very}
Christo Kirov, John Sylak-Glassman, Roger Que, and David Yarowsky. 2016.
\newblock Very-large scale parsing and normalization of wiktionary
  morphological paradigms.
\newblock In \emph{Proceedings of the Tenth International Conference on
  Language Resources and Evaluation (LREC 2016)}, Paris, France. European
  Language Resources Association (ELRA).

\bibitem[{Miller(1998)}]{miller1998wordnet}
George~A Miller. 1998.
\newblock \emph{WordNet: An electronic lexical database}.
\newblock MIT press.

\bibitem[{Ruder et~al.(2018)Ruder, Cotterell, Kementchedjhieva, and
  S{\o}gaard}]{ruder2018discriminative}
Sebastian Ruder, Ryan Cotterell, Yova Kementchedjhieva, and Anders S{\o}gaard.
  2018.
\newblock \href {http://aclweb.org/anthology/D18-1042} {A discriminative
  latent-variable model for bilingual lexicon induction}.
\newblock In \emph{Proceedings of the 2018 Conference on Empirical Methods in
  Natural Language Processing}, pages 458--468. Association for Computational
  Linguistics.

\bibitem[{Sylak-Glassman(2016)}]{sylak2016composition}
John Sylak-Glassman. 2016.
\newblock The composition and use of the universal morphological feature schema
  (unimorph schema).
\newblock Technical report, Technical report, Department of Computer Science,
  Johns Hopkins University.

\bibitem[{Vuli{\'{c}} and Moens(2013)}]{Vulic2013a}
Ivan Vuli{\'{c}} and Marie-Francine Moens. 2013.
\newblock {Cross-Lingual Semantic Similarity of Words as the Similarity of
  Their Semantic Word Responses}.
\newblock In \emph{Proceedings of the 2013 Conference of the North American
  Chapter of the Association for Computational Linguistics: Human Language
  Technologies (NAACL-HLT 2013)}.

\bibitem[{Zhang et~al.(2017)Zhang, Liu, Luan, and Sun}]{zhang2017adversarial}
Meng Zhang, Yang Liu, Huanbo Luan, and Maosong Sun. 2017.
\newblock Adversarial training for unsupervised bilingual lexicon induction.
\newblock In \emph{Proceedings of the 55th Annual Meeting of the Association
  for Computational Linguistics (Volume 1: Long Papers)}, volume~1, pages
  1959--1970.

\bibitem[{Zipf(1949)}]{zipf1949human}
George~Kingsley Zipf. 1949.
\newblock Human behavior and the principle of least effort.

\end{thebibliography}
\bibliographystyle{acl_natbib}

\newpage

\appendix
\clearpage
\newpage

\end{document}